\documentclass[11pt]{article}
\usepackage[]{EACL2023}
\pdfoutput=1

\usepackage[T1]{fontenc}
\usepackage{times}
\usepackage[inline]{enumitem}
\usepackage{latexsym}
\usepackage[utf8]{inputenc}
\usepackage{amssymb ,multirow}
\usepackage{color}
\usepackage{colortbl,tcolorbox}
\usepackage{xcolor}
\usepackage{graphicx,verbatimbox}
\usepackage{acronym} 
\usepackage{amsthm}
\usepackage{tabularx}
\usepackage{multirow}

\theoremstyle{definition}
\newtheorem{definition}{Definition}[section]

\newcounter{todocnt}

\newcommand{\ourmethod}{PREME}

\acrodef{RL}{Reinforcement Learning}
\acrodef{DRL}{Deep Reinforcement Learning}
\acrodef{IR}{Information Retrieval}
\acrodef{MDP}{Markov Decision Process}
\acrodef{NLU}{Natural Language Understanding}
\acrodef{NLP}{Natural Language Processing}
\acrodef{\ourmethod}{Preference-based Meeting Exploration through an Interactive Questionnaire}
\acrodef{CRF}{Conditional Random Field}
\acrodef{QA}{Question Answering}

\usepackage{microtype}

\vspace{-4em}
\title{PREME: Preference-based Meeting Exploration through \\ an Interactive Questionnaire}

\author{Negar Arabzadeh \\
University of Waterloo \\
  \texttt{narabzad@uwaterloo.ca} \\\And
  Julia Kiseleva \\
  Microsoft Research \\\And
  Ali Ahmadvand \\
   Emory University \\\AND
  Yang Liu \\
  Microsoft Research  \\\And
 Ahmed Hassan Awadallah \\
  Microsoft Research\\\And
Ming Zhong\\
  University of Illinois\\\And
Milad Shokouhi \\
  Microsoft Research }
 
\begin{document}
\maketitle

\vspace{-5em}
\begin{abstract}

The recent increase in the volume of online meetings necessitates automated tools for organizing the material, especially when an attendee has missed the discussion and needs assistance in quickly exploring it. 
In this work, we propose a novel end-to-end framework for generating interactive questionnaires for preference-based meeting exploration. As a result, users are supplied with a list of suggested questions reflecting their preferences. 
Since the task is new, we introduce an automatic evaluation strategy by measuring how much the generated questions via questionnaire are answerable to ensure factual correctness and covers the source meeting for the depth of possible exploration.
\end{abstract}

\section{Introduction}
\label{sec:intro}

In recent years, video conferencing technology has gained substantial improvements, and thus, online meetings have become easily accessible and more prominent. Primarily due to the pandemic and working from home, the need for video calling has grown significantly. 
Therefore, the high volume of online meetings necessitates automated tools for managing and organizing essential information for attendees. Especially 
when an attendee has missed an online meeting, it is critical to access the required information since quickly 
reading through the transcript is quite time-consuming. 

\begin{figure}
\centering

\includegraphics[scale=0.4]{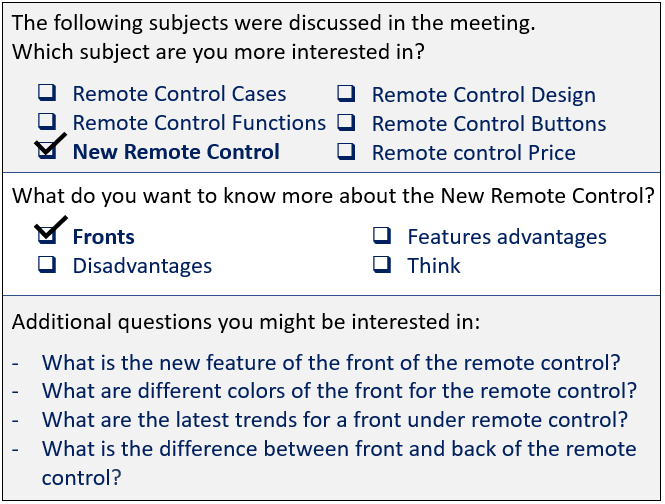}
\vspace{-0.5em}
\caption{An example of exploring one of the meetings from the collection~\citep{carletta2005ami} based on user preferences through an interactive questionnaire. 
}
\vspace{-0.5em}

\label{fig:example}
\vspace*{-5mm}
\end{figure}

\begin{figure*}
\centering
\includegraphics[clip, trim=0.5cm 5.5cm 0.2cm 2cm,scale=0.45]{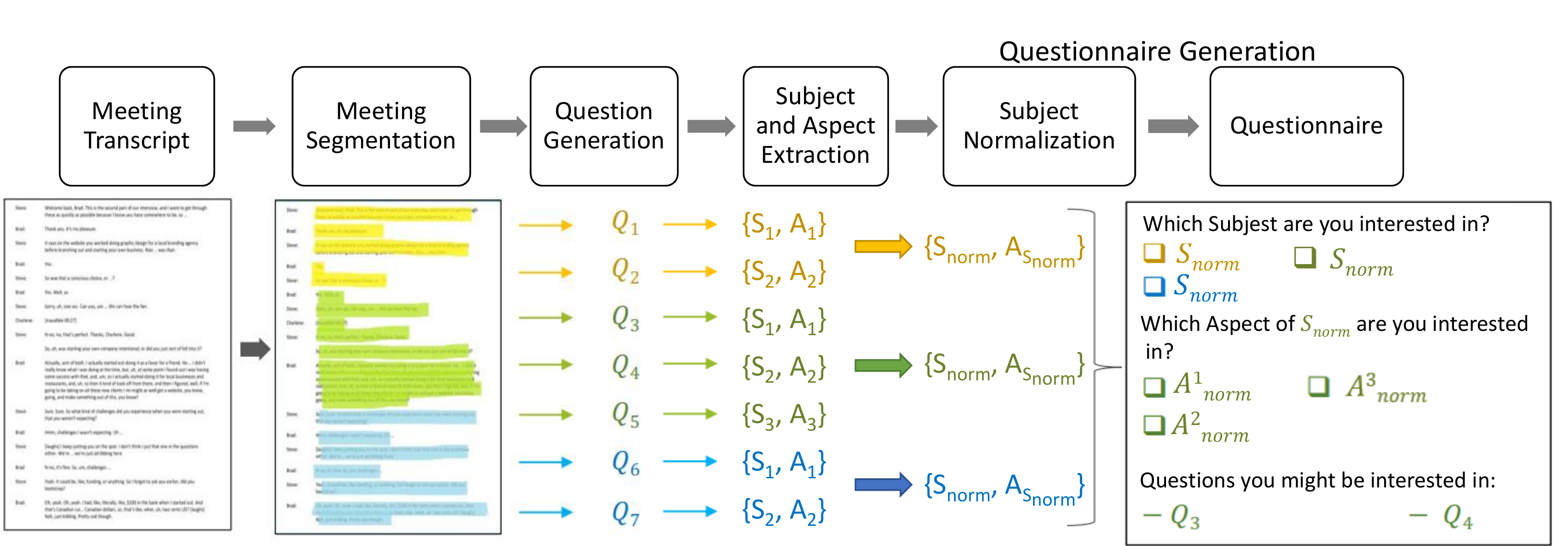}
\vspace{-0.5em}
\caption{Overview of our framework, \acf{\ourmethod}, where $Q$ is a comprehensive set of questions, and $S_i$ and $A_j$ are extracted pairs of subjects and aspects. 
}
\label{fig:flow}
\vspace*{-6mm}
\end{figure*}

Providing meeting summaries is a promising direction~\citep{wang-cardie-2013-domain,jacquenet2019meeting,zhao2019abstractive,singhal2020abstractive}. However, recent studies show that 1) users’ needs do not fully align with current approaches to automatic text summarization \cite{ter2020makes,ter2022summarization} and 2) 
  approaches designed for document summarization could not effectively apply to meetings transcripts  \citep{murray2010generating,mehdad2013abstractive,li-etal-2019-keep} due to the following potential reasons:
\begin{enumerate*}[label=\textbf{(R\arabic*)}]
    \item \textbf{Structure:} standard documents are well structured compared to meeting transcripts;
    \item \textbf{Language:} spoken language used in meetings is less regular than documents; and
    \item \textbf{Multiple speakers:} the speaker role is essential. 
\end{enumerate*}
Moreover, there is little meeting data publicly available that can be used for experimentation compared to regular documents such as news or articles.
In contrast with document summarization, when summarizing a meeting, different users tend different preferences on what content should be included in the summary. Therefore, there is an increasing calling for alternative ways of summarizing, especially for meetings transcripts.
Recently,~\citet{zhong2021qmsum} attempted to tackle this problem by proposing a query-based multi-domain meeting summary, where a user provides a query in question form, e.g., \emph{`What was the discussion about the jog dial's function when talking about changes in the current design?'} to locate the part of the transcript that related to the query and then summarize. However, when attendees have missed the meeting, they cannot formulate such questions due to no prior knowledge about the meeting. To overcome this, we aim to address the following \textbf{research challenge}: \emph{How can attendees effectively explore a meeting content without having prior knowledge about it?}

This work is motivated by the fact that asking questions is a more efficient way for humans to acquire information than notes in plain text~\citep{lawson2007techniques,Lawson2006,aliannejadi-etal-2021-building}. Thus, we address 
preference-based meeting exploration by automatically generating a structured interactive questionnaire for a transcript that covers most of the discussed topics and quickly walks users through the discussed content.
An example of the desired questionnaire is shown in Fig.~\ref{fig:example}. 
First, the user has the ability to express their preferences regarding \emph{subjects} that have been discussed~\citep{solbiati2021unsupervised,huang2018automatic,zhang2019topic,sehikh2017topic}. Next, the questionnaire interactively suggests narrowing down their exploration if possible by displaying a list of possible related \emph{aspects}. As a result, a ranked list of questions reflecting user preferences is generated. Next, the user can pick a question that demonstrates their seeking needs the most and is redirected to the meeting part containing an answer.
Interactively asking for preferences in the  questionnaire  is beneficial because the user oversees what has been covered during the meeting they have missed. In section \ref{survey} we elaborate on  a user study on a number of  professionals who find such application useful for their daily job. 
Hence, the goal of proposed questionnaires is two-fold: 
\begin{enumerate*}[label=\textbf{(G\arabic*)}]
    \item to compactly represent the discussed content; 
    \item to guide users to form questions that express their preference regarding the transcript. 
\end{enumerate*}
We require the generated questionnaire to satisfy the following properties:
\begin{enumerate}[label=\textbf{P\arabic*}, leftmargin=*, nosep]
    \item \label{item:coverage} \emph{Coverage:} coverage is the amount of the information from the source text that a questionnaire points to. The generated questionnaire must cover the meeting as much as possible; 
    \item \label{item:answerable} \emph{Answerable:} a given meeting transcript should contain the answers to the questions generated as a result of the questionnaire. 
\end{enumerate}


To address the defined challenge, we propose a framework, \ac{\ourmethod}, which consists of several concrete sequential steps highlighted in Fig.~\ref{fig:flow}. We start by enchaining the method to extract meeting segments~\citep{solbiati2021unsupervised}. 
Due to the conversational nature of the meeting, topic detection from the segments is challenging~\citep{huang2018automatic,zhang2019topic,sehikh2017topic}. Thus, we indirectly extract the topics as follows. First, we generate questions from each segments~\cite{brown2020language} since extracting topics from the questions is much more well studied. Further, we employ a trained \ac{CRF} model to tag subjects and aspects (Fig.~\ref{fig:example}) from generated questions originated from each segments~\cite{wallach2004conditional}. Once we got each segment's topic list, we proposed a strategy to normalize them to reduce the number of options in the questionnaire.  Recently, ~\citet{DBLP:journals/corr/abs-2010-00490} demonstrated that QA-Based evaluation is strongly correlated with human opinion. Thus, to evaluate \ac{\ourmethod}, we employ a similar QA-based strategy.


To summarize, the main contributions are:
\begin{enumerate}[leftmargin=*,label=\textbf{C\arabic*},nosep]
    \item We propose \ac{\ourmethod}, a novel framework to enable  meetings exploration based on user's preferences through an interactive questionnaire;
    \item We propose a new method for subject normalization which returns the most informative subject from a set of phrases and keywords; 
    \item We introduce a new automatic evaluation strategy for measuring the effectiveness of the proposed questionnaire to assess the required properties \ref{item:coverage} and \ref{item:answerable}, which according to~\cite{DBLP:journals/corr/abs-2010-00490} has a strong correlation with human judgments; and 
    \item We open-source a dataset that includes $1000$ questions comprehensively annotated with subject to their subjects and aspects at \url{https://github.com/microsoft/preme} 
\end{enumerate}
\vspace{-0.5em}
\section{Related Work}
\label{sec:rel_work}
\vspace{-0.5em}
\subsection{Automatic Textual Summarization}
Automatic text summarization task has attracted lots of attention across \ac{NLP} community recently. Many systems are proposed to summarize documents in different domains, including news \citep{rush2015neural,nallapati2017summarunner,DBLP:journals/corr/SeeLM17,celikyilmaz2018deep,liu2019text,zhang2020pegasus}, academic papers~\cite{manakul2021long,huang2021efficient} and books~\cite{kryscinski2021booksum}.
Meeting summarization has also emerged as a widespread need recently.
Due to the unique discourse structure of dialogues, conventional document summarization systems are facing challenges when summarizing meetings~\cite{li-etal-2019-keep,zhu-etal-2020-hierarchical}. 
Thus, new models are proposed for tackling this task.
\citet{wang-cardie-2013-domain} employ decisions, and action items in dialogues to generate the summary progressively.
\citet{oya-etal-2014-template} propose a template-based meeting summarization system by learning the relationship between summaries and their source meeting transcripts.
\citet{shang-etal-2018-unsupervised} design an unsupervised meeting summarization model with multi-sentence compression techniques. 
\citet{li-etal-2019-keep} introduce multi-modal information into meeting summarization with a hierarchical attention mechanism.
\citet{zhu-etal-2020-hierarchical} propose a hierarchical meeting summarizer that can process both word-level and turn-level information of dialogues.
%
Furthermore, the community noted that due to the lengthy content and distributed information, a  general summary of the meetings does not necessarily satisfy what users seek.
Thus, Query-based summarization methods have become more prevailing for generating concise and specific summaries. \cite{litvak-vanetik-2017-query,nema-etal-2017-diversity,baumel2018query,ishigaki2020neural,kulkarni2020aquamuse,kulkarni2021comsum,pasunuru2021data}. 
Recently, \citet{zhong2021qmsum} proposed a new framework of query-based summarization for meetings, in which they annotate QMSUM, a query-based multi-domain meeting dataset.
Each QMSUM meetings come along with a set of queries with different levels of abstractness, i.e., general queries and specific queries.
Human annotators write these queries, and the summaries align with these queries after reading the meeting transcripts.

While query-based summarization can be a proper path to provide users with meeting information at different specificity levels, we argue that issuing such specific queries still requires a certain degree of background knowledge. 
In real-life scenarios, users might not be equipped with that knowledge and issue informative queries, especially when they did not attend the meeting. 
Hence, they can not benefit from query-based summarization techniques to explore the meetings. We address the drawbacks of query-based summarizers by providing users with an interactive questionnaire which provides them with potential queries and allows them to explore the meetings more flexibly.

\subsection{Evaluation of Summaries Factuality}

The summaries often has called out for hallucination issues~\citep{maynez-etal-2020-faithfulness}.
Thus, \citet{wang-etal-2020-asking} propose a framework to evaluate factual consistency of summaries with the source text 
Similarly, \citet{DBLP:journals/corr/abs-2010-00490} propose a \ac{QA}-based evaluation approach on summaries' content quality. They measure how much information is contained in a candidate summary by calculating the proportion of questions it can answer.
These approaches inspired us for automated end-to-end evaluations of the  questionnaires.
\vspace{-0.5em}
\subsection{Question Generation and Filtering}
Initial works in Question Generation task leveraged crowd-sourcing or rule-based methods to generate pre-defined question templates~\citep{mostow2009generating,rus2010first,lindberg2013generating,DBLP:journals/corr/abs-2004-11892,mazidi-nielsen-2014-linguistic,labutov2015deep}. \citet{heilman2010good} tackled this problem by over-generating candidate questions and then using a learning to rank framework to rank them to filter the low-quality questions.
SQUASH~\citep{krishna2019generating} is one of the recent works in which authors used question generation methods to convert a document into a hierarchy of question-answer pairs with the focus on questions' granularity level. 
They employed a neural encoder-decoder model trained on three reading comprehension data sets, i.e., SQuAD~\cite{rajpurkar2016squad}, QuAC~\cite{choi2018quac}, and CoQ~\cite{reddy2019coqa} to generate the questions, and further, they filtered out the unanswerable questions using some heuristics and question answering models. While question generation using question answering data sets seems a general approach, this method does not work well on meeting-related questions generated due to many reasons, including:
\begin{enumerate*}[label=\textbf{(\arabic*)}]
    \item Different structure of meetings compared to documents;
    \item There are not many question-answering datasets available from meetings;
    \item Sometimes, the answer to questions generated from meetings could be very long, making it hard to fit the context in neural models.
\end{enumerate*}
In our work, we introduce an automatic method that can generate questions regarding the meeting to overcome the high price of collecting with annotators.
\vspace{-0.5em}
\subsection{Questionnaire Organization}
Obtaining users preferences has always shown to be a challenging task ~\citep{jiang2008mining,rokach2012initial,anava2015budget,christakopoulou2016towards,sepliarskaia2018preference}. The task becomes more challenging when we aim to minimize the number of interactions with users to get to know their preferences.  \citet{sepliarskaia2018preference} reformulate this task as an optimization problem. They propose a static questionnaire by choosing a minimal and diverse set of questions. 
Similarly, in \citet{liu2019dquest} proposed a dynamic questionnaire generation method for search of clinical trials. 
Quiz-style question generation has also been explored recently by \citet{lelkes2021quiz}. The authors have formulated the problem as two sequence to sequence tasks, including the question-answer generation step and incorrect answer generation step. We argue that while the former step seems relevant to our work, it could not be adapted to meeting transcripts since their proposed dataset has been trained on factual question answering data sets and cannot be used for meeting purposes. All in all, we can conclude that creating questionnaires are still under exploration in different domain. Hence, our effort in organizing a questionnaire, especially for meetings, is timely and useful for future research.
\vspace{-0.33em}
\section{Proposed Framework: \ac{\ourmethod}}
\label{sec:approach}

This section explains \ac{\ourmethod}, our proposed novel methodology to explore meetings based on users' preferences through an interactive questionnaire. An overview of our methodology is shown in Fig.~\ref{fig:flow} in which we first apply a topic segmentation method 
~\cite{solbiati2021unsupervised} on meeting transcript to retrieve segments with different topics (Section~\ref{sec:meeting_segmentation}). Then, we generate a set of all possible questions from each segment (Section~\ref{sec:qg}). Further, we extract the most informative part of the questions, i.e., the subject and aspect of each question (Section~\ref{sec:topic_aspects}). In the last step, we map the normalized subjects and aspects with generated questions and form the questionnaire (Section~\ref{sec:uestionnaire}).

\subsection{Meeting Segmentation}
\label{sec:meeting_segmentation}

A meeting transcript can be extremely long and contain discussions of various topics.
 Therefore, our goal is to divide the meeting text into a sequence of topically coherent chunks. 
Thus, we adopted an unsupervised topic segmentation method based on the contextualized presentation of meeting~\citep{solbiati2021unsupervised}. In this topic segmentation method, 
the authors compute the BERT embeddings for every utterance of the meeting transcript. Further, they curated blocks of utterances and performed a block-wise max-pooling operation to generate contextualized embedding for each block. Then, the semantic similarity between two adjacent blocks is captured, and a change in the topic is detected if two adjacent blocks show similarity below a certain threshold. 
This approach has several advantages, including:
\begin{enumerate*}[label=\textbf{(\arabic*)}]
    \item It is unsupervised;
    \item Since 
    we are just converting the meeting into smaller pieces, and we are not losing any part of the meeting. 
\end{enumerate*}

\vspace{-0.5em}
\subsection{Question Generation}
\label{sec:qg}
For question generation from a segment, we leveraged the powerful GPT-3 model \cite{brown2020language}.
%
%
An impressive capability of the GPT-3 is to generate very realistic results from few training samples or even no training sample (few-shot and zero-shot learning). The variety of the generated content can be controlled using a temperature hyper-parameter. To expand the size of generated questions' pool as much as possible,  in each segment, the API is called in a zero shot learning model with different temperature values between [0-1] with a 0.05 margin, where the value closer to 1 means more diversified questions. We set the maximum output length to 128 tokens and then we repeat the process for 10 trials for each specific temperature. Given that the  maximum context window for the API was 2048 tokens, we truncate and slide  by half-a-window size of 2048 tokens whenever a segment includes more than 2048 tokens. As a results, A list of questions is extracted based on random initialization in each API call, meaning different results are achieved even with the same hyper-parameters. We extracted five questions on average per segment in each call. Finally, a union across all runs is used to form our question pool.

\begin{table}[]

\caption{Examples of annotated questions with their subjects and aspects 
.
\colorbox{red!40}{Subjects} are highlighted in red and \colorbox{green!40}{Aspects} are highlighted in green.}
\vspace{-1em}
\scalebox{0.74}{

\begin{tabular}{|l|p{8.5cm}|}
\hline 
\rowcolor[HTML]{E8E8E8} Q1 & What is the \colorbox{green!40}{arrow symbol} on the  \colorbox{red!40}{remote control} for? \\[9pt]
Q2 &What are the \colorbox{green!40}{main frustrations people have} with the \colorbox{red!40}{remote control}? \\[20pt]
\rowcolor[HTML]{E8E8E8}  Q3 & How will the \colorbox{green!40}{logo and color scheme} be incorporated into the \colorbox{red!40}{product}? \\[20pt]
Q4 & What are \colorbox{green!40}{pros and cons} of having a \colorbox{red!40}{remote with a large number of buttons}? \\[20pt]
\rowcolor[HTML]{E8E8E8}  Q5 & What is the \colorbox{red!40}{most difficult part of the project} \colorbox{green!40}{from the industrial engineer's point of view}? \\[20pt] \hline
\end{tabular}}
\vspace{-1.2em}
\label{examples}
\end{table}

\vspace{-0.5em}
\subsection{Subject and Aspect Extraction}
\label{sec:topic_aspects}

Every of the generated questions has one or more \emph{subject(s)} that is defined as the principal matter that attendees have discussed, i.e., the main concern of the questions. Some questions might point to  a specific \emph{aspect(s)} of the subject which is defined as the mentioned details about a given subject. We aim to extract the primary subjects from any question and the detailed aspect if it is mentioned. Table~\ref{examples} shows examples of annotated \emph{subjects} and \emph{aspects} for a few questions. For instance, in the question \emph{``What is the arrow symbol on the remote control for?"}, ``remote control" is annotated as the subject and the ``arrow symbol" is the specific  aspect of the subject.  
To extract the subjects and aspects from the questions, we use \ac{CRF}~\cite{wallach2004conditional}. We examined SOTA keyword extraction and contextualized neural embedding-based topic extraction models; however, the CRF model which uses word's identity, suffix, shape and POS tags as features,
seems to work the best among them. To train the \ac{CRF} model, we were required to have annotated questions with subjects and aspects labels. We designed an annotation study using the UHRS\footnote{\url{https://prod.uhrs.playmsn.com/uhrs/}} crowd-sourcing platform, where we carefully trained annotators with detailed instructions to label randomly selected 1000 questions generated by GPT3 with their subject and aspects\footnote{We invested in having a few well-trained annotators rather than having a high number of annotators who have not been trained well. Thus, annotators were paid hourly and by the quality of their work and they had no intentions for cheating.}. Each question has been assigned to two annotators, and we report the annotators' agreement  in Section~\ref{sec:eval}. Further, we employ the trained CRF model to extract subjects and aspects from the questions.

\subsection{Questionnaire Generation}
\label{sec:uestionnaire}
Given a meeting transcript, for each of its segment $T$  which was initially supposed to coherently point out one subject, we generate $Q_T$, a set of generated questions from $T$. In other words, given an ideal meeting segmentation method, each segment is supposed to be pointed to one subject. Thus, we assume that each segment has only one valid topic and as shown in Figure \ref{fig:flow}, each segment is being represented with one $S_{norm}$.
We create a set $S_{Q_T}$ by extracting the subjects from each question in $Q_T$. Therefore, for the segment $T$, we have at least $|Q_T|$ number of subjects. Extracted subjects from a question set with the same origin segment must be normalized so that one comprehensive, general, and informative subject presents a segment. The more the selected subject representative covers other concepts in $S_{Q_T}$, the better normalization we employed. This subject normalization reduces the number of subjects shown to the user at the first step of the questionnaire and will decrease the user's effort, 
causing figuring out users' preferences by asking them the minimum number of questions. In other words, our goal is to select a single subject $S_{norm}$ from $S_{Q_T}$ which represents $S_{Q_T}$in the most informative way. To do so, we define the notion of the subject network as follows. 


\begin{figure*}
\centering
\vspace{-0.5em}
\includegraphics[scale=0.5]{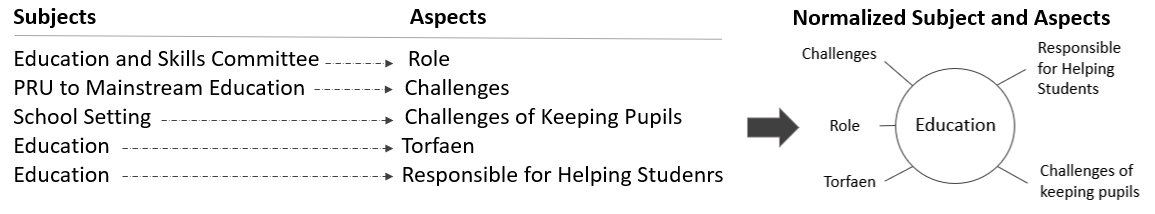}
\vspace{-2.5em}
\caption{An example of how extracted subjects and aspects from a given segment are normalized.}
\label{fig:norm}
\end{figure*}

\begin{figure}
\centering
\vspace{-2em}
\includegraphics[scale=0.5]{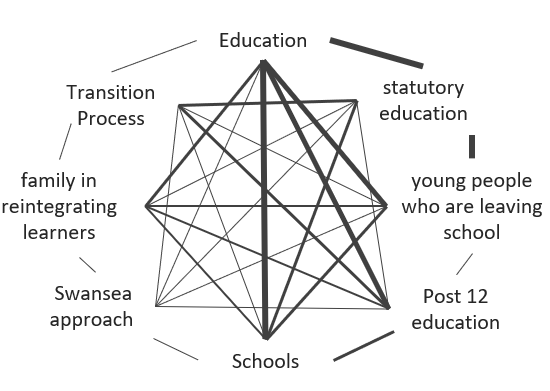}
\vspace{-1em}
\caption{An example of subject-network built for one extracted segments from \cite{janin2003icsi}.  The edge weights represent the semantic similarity between each nodes. Higher weights are shown with higher width.
}
\vspace{-1.5em}
\label{fig:topics}
\end{figure}

\begin{definition}
\label{ego_def}
Given a segment $T$, a set of generated questions $Q_T$, and extracted subjects $S_{Q_T}$,  a subject-network for $G(S_{Q_T})$ is denoted as  $G(S_{Q_T})= (\mathbb{V}, \mathbb{E}, \textit{w})$. It is a weighted undirected graph, where $\mathbb{V}= \{ s_i \in S_{Q_T} \}$, and $\mathbb{E} = \{ e_{s_i} ,  e_{s_j} : \forall s_i , s_j \in	\mathbb{V} \}$. The function $\textit{w} :\mathbb{E} \to [0, 1]$ is the  cosine similarity between the semantic relatedness of the contextualized embedding vectors of two incident subjects of an edge $e_{s_i,  s_j}$, i.e., $v_{s_i}$ and $v_{s_j}$. 
\end{definition}


\begin{figure*}
\centering
\includegraphics[clip, trim=4.1cm 6.4cm 6.5cm 6cm,scale=0.65]{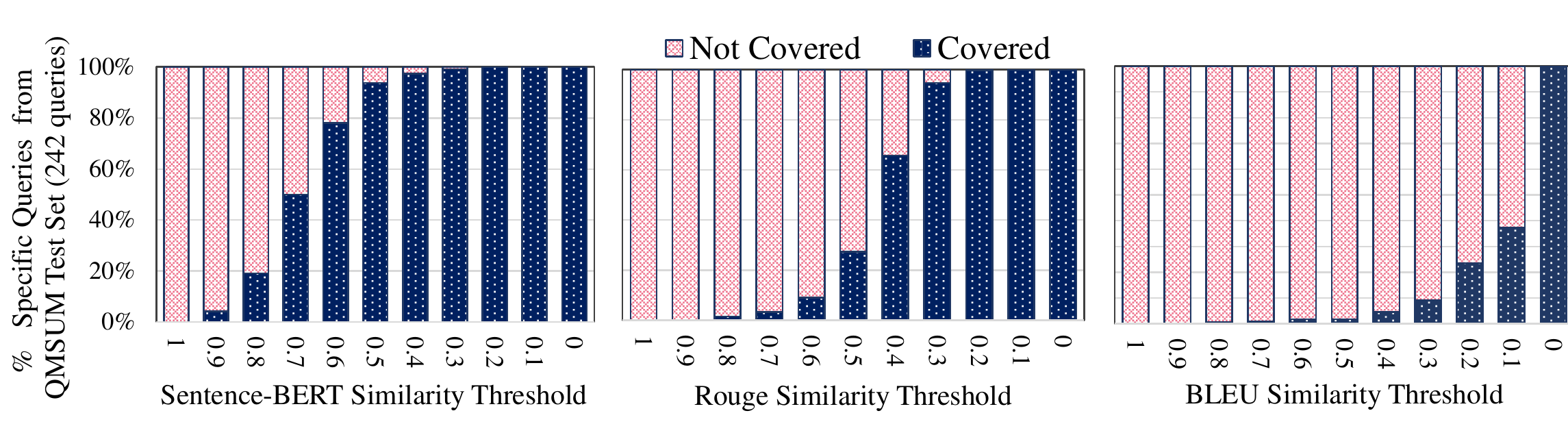}
\vspace{-1.5em}
\caption{Coverage of \ourmethod on QMSUM test set  considering different similarity metrics and threshold 
}
\vspace{-1em}
\label{fig:similarity}
\end{figure*}

In Def.~\ref{ego_def}, we propose a subject-network  
where subjects are connected, and edge weights represent the semantic similarity between the two subjects.
We hypothesize that the node with  highest  similarity and connection to others is the most central one. In other words, since it has great similarity to other subjects, there is a high probability that it points to a more generic concept and that covers the other subjects. Hence, the node $S_{norm}$ should have high centrality attribute to represent the main subject of segment $S$. We employed PageRank~\citep{haveliwala2003topic} value to find the most important and informative node in this network. Similarly, PageRank has shown to have a high correlation with the most important nodes and has been used in tackling different tasks such as quantifying term's specificity or ranking problems in different information retrieval tasks \cite{arabzadeh2020neural,arabzadeh2019geometric,kurland2010pagerank}. We measure the PageRank score of each node and select the node with the highest PageRank value as the representative subject $S_{norm}$ of the subject set $S_{Q_T}$ for segment $T$. In other words, we represent each segment $T$ by subject $S_{norm}$ where $PageRank(S_{norm}$) $>$ $PageRank$ ($s_i$) for every $s_i \in \mathbb{V}$.

Fig.~\ref{fig:topics} displays a subject-network generated from extracted subjects from one of the meetings' segments in the QMSUM dataset. subjects such as ``Education", ``Schools," ``Young people who are leaving school" are included in this subject set and represented by nodes in this subject-network. Further, we connect every pair of nodes in this graph, and the edge weight is directly related to their semantic similarity. As presented in Fig.~ \ref{fig:topics}, some nodes have higher edge weights which their connected lines are shown with greater width. We measure page rank in this weighted network. Here ``Education" got the highest PageRank value in this subject-network. Hence, we present these subjects by one subject, i.e., ``Education". ``Education" can be a promising representative for these subjects as it covers more specific concepts such as ``schools", ``statutory education," and ``post 12 education."

Next, the extracted aspects from each question set should be mapped to their representative subject. We remove the redundant and repetitive aspects and subjects by removing those who have highly similar n-grams. Plus, There might be several subjects existing in $S_{Q_T}$ which all point out to $S_{norm}$, and they might be semantically very similar.  In this step, we must be concerned not to lose any aspect because of subject normalization. We aim to map every aspect from $S_{norm}$ and every $s_i$ in $S_{Q_T}$ which is highly similar to $S_{norm}$ to maximize the potential of questions we might want to show at the end of the questionnaire. For instance, in Fig~\ref{fig:norm}  we display a few extracted subjects and aspects from one segment. If we only consider ``education" and its related aspect, we will lose many aspects that users might be interested in, and as a result, the questionnaire coverage will drop.
On the other hand, if we merge the highly similar representative subjects with, e.g., ``school setting" and ``Education and Skills Committee," we will have a broader host of questions to suggest to users. Therefore, we will filter out dissimilar subjects from $S_{Q_T}$ to $S_{norm}$ and  map extracted aspects from filtered $S_{Q_T}$ to $S_{norm}$ as it is shown in Fig.~\ref{fig:norm}. As a result, if ``education" is the subject of interest for a user, they have the opportunity to select which aspects of education they are more interested in, such as "Role" of education or ``challenges" of education. Finally, we will show users the questions in which the selected aspects and normalized subjects have appeared. 

\vspace{-0.5em}
\section{Evaluation Methodology}
\label{sec:eval}

For experiments, we use the QMSUM dataset~\cite{zhong2021qmsum}, which includes $232$ product, academic, and committee meetings ~\citep{janin2003icsi,carletta2005ami}. 
 Each meeting comes with a set of general and specific questions; the general ones are out of the scope of this work since they refer to very broad concepts, e.g., \emph{``summarize the whole meeting."}. 
Further evaluations are conducted on the QMSUM test set.

\begin{table}[]
\vspace{-1em}
\caption{Annotators agreement on annotated questions with respect to subjects and aspects using Kripendorff's score~\cite{krippendorff2011computing}}
\vspace{-1em}
\centering
\resizebox{0.99\columnwidth}{!}{
    \begin{tabular}{lll}
     & \bf Subject & \bf Aspect \\ 
    \bf Hard [Exact Match] & 0.459 & 0.415 \\ 
    \bf Soft [At least one term matched] & 0.490 & 0.485 \\ 
\end{tabular}
}
\vspace{-0.5em}
\label{tab:agg}
\vspace*{-3mm}
\end{table}

\vspace{-0.5em}
\subsection{Evaluating Framework Components}

The proposed framework consists of several steps (Fig.~\ref{fig:flow}). The used \emph{meeting segmentation}~\citep{solbiati2021unsupervised} method  has shown to outperform baselines ~\citep{hearst1997text,beeferman1999statistical,badjatiya2018attention}. Hence, we refer to original paper for evaluation results. 

\textbf{Evaluating Question Generation:}
We evaluate the quality of our generated questions by measuring the fraction of generated questions by human annotators in QMSUM that we covered in \ac{\ourmethod}. We assume the specific queries in the QMSUM dataset enjoy relatively high quality because annotators issued them after comprehensively reading the transcript (gold standard questions). Hence, Fig.~\ref{fig:similarity} reports the similarity between most similar questions generated by \ac{\ourmethod} and the gold questions by three different similarity metrics i.e., Sentence-BERT similarity~\citep{DBLP:journals/corr/abs-1908-10084}, Rouge F-1 score~\citep{lin-2004-rouge}, and BLEU-4 score~\citep{papineni2002bleu}. We assume a questions from QMSUM is covered  if there is at least a question generated by \ac{\ourmethod} that has similarity is higher than a certain threshold $t \in [1,0.9,...,0.1,0]$. We report the percentage of `Covered/Not Covered' questions based on different similarity matching thresholds. Based on Fig.~\ref{fig:similarity} we conclude while we cover a relatively fair number of specific questions, there is still room for improvement. However, we should note that the questions in QMSUM are very limited, and initially, they were not supposed to cover all possible questions that one could raise from the meeting. Additionally, we observe that questions in QMSUM, which are issued by humans, include more abstractive questions while our generated questions inclined toward more factual ones.

\begin{table}
\caption{CRF performance on extracting subjects and aspects of questions using 10-fold cross validation 
}
\vspace{-0.5em}
\centering
\resizebox{0.75\columnwidth}{!}{
  \begin{tabular}{llll}
     & \bf Precision & \bf Recall & \bf F1-Score \\ 
    \bf Subject & 0.64 & 0.69 & 0.67 \\ 
    \bf Aspect  & 0.89 & 0.80 & 0.84 \\ 
    \bf N/A     & 0.63 & 0.73 & 0.68 \\
 \end{tabular}
}
\label{tab:CRF-performance}
\vspace{-2em}
\end{table}

\textbf{Evaluating Subject and Aspect Extraction:}
To assess the quality of the collected dataset, we measure Krippendorff's alpha agreement between annotators~\cite{krippendorff2011computing} for extracted \emph{subject} and \emph{aspect} of the $1000$ questions generated from the training set. Tab.~\ref{tab:agg} shows annotators have agreement $\sim 0.4$, which is interpreted as ``Moderate'' agreement for such a challenging task. Since different annotators might selected different section of the text, Tab.~\ref{tab:agg} reports both \emph{hard} and \emph{soft} agreements.
%
we trained the \ac{CRF} model using \textit{crfsuite} library and evaluated it by 10-fold cross-validation. Given each term in the questions, the model predicts whether the term is considered the subject, aspect, or not applicable for labeling (N/A). Tab.~\ref{tab:CRF-performance} shows the result of the \ac{CRF} model evaluation in terms of precision, recall, and F1 scores. We notice that the model shows better performance on detecting aspects compared to the subject. 

\vspace{-0.5em}
\subsection{Evaluating Questionnaires}
\label{survey}

To the best of our knowledge, we are first to propose a preference-based questionnaire as a way for meeting exploration; thus, no particular gold standard benchmark or evaluation metrics. Since we require users to express their preference, it makes it challenging to simulate \emph{`enough imaginative context'} among annotators.
Thus, we conducted a user study to highlight the usefulness of exploring meetings through an interactive questionnaire. 
We provided 20 participants who were professional workers and graduate students aged between 24-41  with detailed explanations and examples of results generated by PREME such as in Figure \ref{fig:example}. Participants on average had over 5 hours of online meetings per week. Among which, over 80\% of them reported that they need to explore the content of a past meeting, at least a couple of times a week.  Finally, over 80\% of participants agreed on finding PREME useful for meetings exploration.
Also, we introduce a new evaluation strategy that satisfies the desired properties on coverage~(\ref{item:coverage}) and the existence of answers in the transcript~(\ref{item:answerable}). 
The proposed automatic metrics capture if our framework is ready to be tested through a more comprehensive user study in the future, when we can run a pair-wise preference-based comparison between PREME and other meeting exploration methods.

\begin{table}[t]
\caption{Test set statistics and \ac{\ourmethod} Performance: Average number of generated questions and Coverage.}
\vspace{-0.5em}
\label{tab:coverage}
\centering
\resizebox{1.0\columnwidth}{!}{
    \begin{tabular}{lllll}
     & \bf \#Meetings & \begin{tabular}[c]{@{}l@{}}\bf Average \# \\ \bf Turns \end{tabular} & \begin{tabular}[c]{@{}l@{}}\bf Average \# \\ \bf Questions\end{tabular} & \bf Coverage (\%) \\ 
    \bf Academic & 9 & 893 & 1257 &  83.07\%\\ 
    \bf Committee & 6 & 214 & 1105 &  64.04\%\\ 
    \bf Product & 20 & 569 & 724 &  86.25\%\\ 
    \bf All & 35 & 591 & 927 &  81.62\%\\ 
    \end{tabular}
}
\vspace{-1em}
\end{table}

\textbf{Automatic evaluation}: We utilize the model SOTA called Locator in~\citep{zhong2021qmsum} in which, given the query, it can extract the relevant spans from the meeting. The Locator employs a hierarchical ranking-based model structure based on CNN~\citep{DBLP:journals/corr/Kim14f} and Transformers~\citep{vaswani2017attention} architecture. The Locator embeds each utterance of the meeting and feeds it to a CNN network by capturing the local features, and utilize Transformer layers to obtain contextualized turn-level representations. In addition, the speaker's embedding is also concatenated to the features list. Finally, the model uses MLP to score each turn, and the turns with the highest scores are considered the relevant spans for each question. 

\begin{figure}
\centering
\vspace{-0.5em}
\includegraphics[clip, trim=11cm 6.8cm 7cm 6.5cm,scale=0.66]{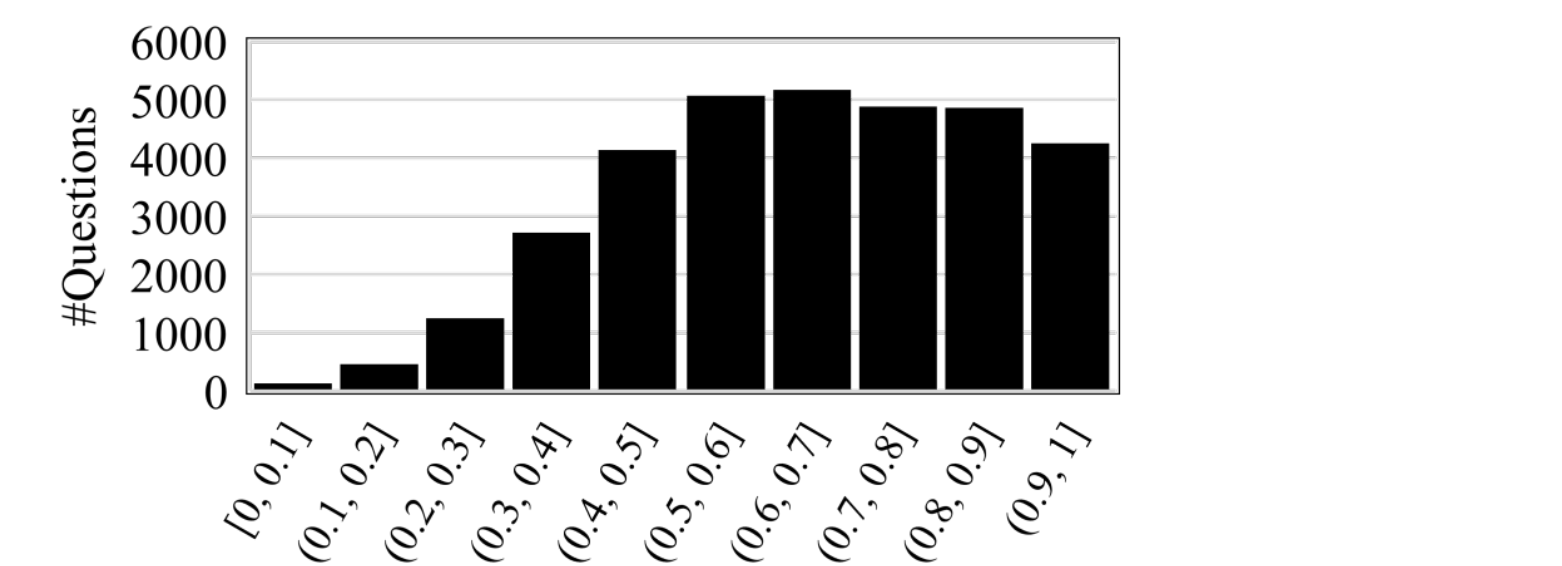}
\caption{Histogram of Confidence Scores of Question-Answering model on generated questions from \ac{\ourmethod}.}
\vspace{-1.5em}
\label{fig:confidence}
\end{figure}

\vspace{-0.3em}
To measure the coverage (to satisfy~\ref{item:coverage}), we adopt the newly proposed \ac{QA}-style of evaluation~\citep{DBLP:journals/corr/abs-2010-00490, wang-etal-2020-asking} which has shown to have substantial correlation with human judgments in terms of questions quality assessments. \textit{Coverage} is defined as the fraction of a meeting that a questionnaire encompasses. To measure the coverage, first, the relevant answer spans for the existing questions in a questionnaire are located. Further, the proportion of utterances that were already located as relevance answer spans  w.r.t. the whole meeting transcripts, is measured as the coverage. We believe that that is a promising indicator of questionnaire informativeness. The coverage is basically how much of the original meeting was covered by the questionnaire. We hypothesize that a good questionnaire should ideally include questions from all parts of a meeting. i.e., the questionnaire includes questions related to every part of the meeting so that users are able to explore their section of interest from the meeting. Therefore, the more the questionnaire covers the meeting, the better it is. To do so, we find the answer spans to the generated question in each questionnaire and we report the percentage of utterances that the locator detected as the answer span for all the questions in the questionnaire from the whole meeting. 
We run our experiments on the QMSUM test set.  Tab.~\ref{tab:coverage} shows the details of this test set.  
We over generate the questions and after removing the duplicates, on average, the questionnaire has 1257 unique questions from Academic meetings, 1105 questions from Committee meetings, and 724 questions from Product meetings. 
Further, Tab.~\ref{tab:coverage} reports the percentage of utterances covered in each meeting. On average, our proposed questionnaire can cover 81\% of the meeting.
We also compared the coverage on different types of meetings. While our generated questionnaire covered Committee meetings the least (64\%), the Product and Academic meetings show higher coverage (over 80\%). 
Further, we evaluate how much the generated questions in \ac{\ourmethod} are answerable (to satisfy~\ref{item:answerable}). Inspired by~\citep{krishna2019generating}, we run a pretrained \ac{QA} model~\citep{sanh2019distilbert} over generated questions and report the confidence score for each \ac{QA} pair in Fig.~\ref{fig:confidence}. We use DistilBERT fine-tuned on SQUAD~\citep{rajpurkar2016squad} dataset. We observe that more than $73\%$ of generated questions from \ac{\ourmethod} on meetings in test set of QMSUM shows confidence score higher than $0.5$ and more than $42\%$ of questions shows confidence score greater than $0.7$. The results confirm that a promising portion of generated questions are answerable.

\vspace{-0.5em}
\section{Conclusions and Future Work}
\vspace{-0.5em}
\label{sec:conclusions}
We proposed an end-to-end framework, called \ac{\ourmethod}, that allows automatically building a questionnaire that will enable users to explore the most of discussed subjects and their aspects if desired. As a result, users are supplied with 
questions about the meetings that express their information needs, and answers can be found in the transcript. Since simulating actual users' preferences is challenging and requires hired annotators, we have ran a small user study as well running an automatic end-to-end evaluation strategy to demonstrate the desired properties (\ref{item:coverage} and \ref{item:answerable}) of the generated questionnaires.  
We publicly release the collected dataset of annotated questions concerning its subjects and aspects, the code for questionnaires generation, and our evaluation procedure to carry forward the proposed state-of-the-art for the newly formulated problem. In future, and by proposing a new method for questionnaire generation will allow us to run a user study for pair-wise comparison of the methods and reveal the correlation between human and automatic evaluation metrics for the suggested task. 
\section{Limitations}
Generally, there is not much data available for meeting exploration. Thus, all studies on this domain are limited by small training and exploratory data. Therefore, it would be beneficial for the community to collect more labelled meeting data for meeting exploration and organization purposes. Since PREME is made of different SOTA components, its performance is also limited by individual components. In future, novel attempts can be made to address this problem as an end-to-end framework. In addition, the future works should include an extensive human evaluation that will reveal additional requirements for the \ac{\ourmethod} to satisfy, which will suggest additional evaluation metrics. Plus, since this the first work on to tackle meeting exploration via questionnaire, the preference-based evaluation is not possible.

\bibliography{acl}

\begin{thebibliography}{75}
\expandafter\ifx\csname natexlab\endcsname\relax\def\natexlab#1{#1}\fi

\bibitem[{Aliannejadi et~al.(2021)Aliannejadi, Kiseleva, Chuklin, Dalton, and
  Burtsev}]{aliannejadi-etal-2021-building}
Mohammad Aliannejadi, Julia Kiseleva, Aleksandr Chuklin, Jeff Dalton, and
  Mikhail Burtsev. 2021.
\newblock \href {https://doi.org/10.18653/v1/2021.emnlp-main.367} {Building and
  evaluating open-domain dialogue corpora with clarifying questions}.
\newblock In \emph{Proceedings of the 2021 Conference on Empirical Methods in
  Natural Language Processing}, pages 4473--4484, Online and Punta Cana,
  Dominican Republic. Association for Computational Linguistics.

\bibitem[{Anava et~al.(2015)Anava, Golan, Golbandi, Karnin, Lempel, Rokhlenko,
  and Somekh}]{anava2015budget}
Oren Anava, Shahar Golan, Nadav Golbandi, Zohar Karnin, Ronny Lempel, Oleg
  Rokhlenko, and Oren Somekh. 2015.
\newblock Budget-constrained item cold-start handling in collaborative
  filtering recommenders via optimal design.
\newblock In \emph{Proceedings of the 24th international conference on world
  wide web}, pages 45--54.

\bibitem[{Arabzadeh et~al.(2020)Arabzadeh, Zarrinkalam, Jovanovic, Al-Obeidat,
  and Bagheri}]{arabzadeh2020neural}
Negar Arabzadeh, Fattane Zarrinkalam, Jelena Jovanovic, Feras Al-Obeidat, and
  Ebrahim Bagheri. 2020.
\newblock Neural embedding-based specificity metrics for pre-retrieval query
  performance prediction.
\newblock \emph{Information Processing \& Management}, 57(4):102248.

\bibitem[{Arabzadeh et~al.(2019)Arabzadeh, Zarrinkalam, Jovanovic, and
  Bagheri}]{arabzadeh2019geometric}
Negar Arabzadeh, Fattaneh Zarrinkalam, Jelena Jovanovic, and Ebrahim Bagheri.
  2019.
\newblock Geometric estimation of specificity within embedding spaces.
\newblock In \emph{Proceedings of the 28th ACM International Conference on
  Information and Knowledge Management}, pages 2109--2112.

\bibitem[{Badjatiya et~al.(2018)Badjatiya, Kurisinkel, Gupta, and
  Varma}]{badjatiya2018attention}
Pinkesh Badjatiya, Litton~J Kurisinkel, Manish Gupta, and Vasudeva Varma. 2018.
\newblock Attention-based neural text segmentation.
\newblock In \emph{European Conference on Information Retrieval}, pages
  180--193. Springer.

\bibitem[{Baumel et~al.(2018)Baumel, Eyal, and Elhadad}]{baumel2018query}
Tal Baumel, Matan Eyal, and Michael Elhadad. 2018.
\newblock Query focused abstractive summarization: Incorporating query
  relevance, multi-document coverage, and summary length constraints into
  seq2seq models.
\newblock \emph{arXiv preprint arXiv:1801.07704}.

\bibitem[{Beeferman et~al.(1999)Beeferman, Berger, and
  Lafferty}]{beeferman1999statistical}
Doug Beeferman, Adam Berger, and John Lafferty. 1999.
\newblock Statistical models for text segmentation.
\newblock \emph{Machine learning}, 34(1):177--210.

\bibitem[{Brown et~al.(2020)Brown, Mann, Ryder, Subbiah, Kaplan, Dhariwal,
  Neelakantan, Shyam, Sastry, Askell et~al.}]{brown2020language}
Tom~B Brown, Benjamin Mann, Nick Ryder, Melanie Subbiah, Jared Kaplan, Prafulla
  Dhariwal, Arvind Neelakantan, Pranav Shyam, Girish Sastry, Amanda Askell,
  et~al. 2020.
\newblock Language models are few-shot learners.
\newblock \emph{arXiv preprint arXiv:2005.14165}.

\bibitem[{Carletta et~al.(2005)Carletta, Ashby, Bourban, Flynn, Guillemot,
  Hain, Kadlec, Karaiskos, Kraaij, Kronenthal et~al.}]{carletta2005ami}
Jean Carletta, Simone Ashby, Sebastien Bourban, Mike Flynn, Mael Guillemot,
  Thomas Hain, Jaroslav Kadlec, Vasilis Karaiskos, Wessel Kraaij, Melissa
  Kronenthal, et~al. 2005.
\newblock The ami meeting corpus: A pre-announcement.
\newblock In \emph{International workshop on machine learning for multimodal
  interaction}, pages 28--39. Springer.

\bibitem[{Celikyilmaz et~al.(2018)Celikyilmaz, Bosselut, He, and
  Choi}]{celikyilmaz2018deep}
Asli Celikyilmaz, Antoine Bosselut, Xiaodong He, and Yejin Choi. 2018.
\newblock Deep communicating agents for abstractive summarization.
\newblock \emph{arXiv preprint arXiv:1803.10357}.

\bibitem[{Choi et~al.(2018)Choi, He, Iyyer, Yatskar, Yih, Choi, Liang, and
  Zettlemoyer}]{choi2018quac}
Eunsol Choi, He~He, Mohit Iyyer, Mark Yatskar, Wen-tau Yih, Yejin Choi, Percy
  Liang, and Luke Zettlemoyer. 2018.
\newblock Quac: Question answering in context.
\newblock \emph{arXiv preprint arXiv:1808.07036}.

\bibitem[{Christakopoulou et~al.(2016)Christakopoulou, Radlinski, and
  Hofmann}]{christakopoulou2016towards}
Konstantina Christakopoulou, Filip Radlinski, and Katja Hofmann. 2016.
\newblock Towards conversational recommender systems.
\newblock In \emph{Proceedings of the 22nd ACM SIGKDD international conference
  on knowledge discovery and data mining}, pages 815--824.

\bibitem[{Deutsch et~al.(2020)Deutsch, Bedrax{-}Weiss, and
  Roth}]{DBLP:journals/corr/abs-2010-00490}
Daniel Deutsch, Tania Bedrax{-}Weiss, and Dan Roth. 2020.
\newblock \href {http://arxiv.org/abs/2010.00490} {Towards question-answering
  as an automatic metric for evaluating the content quality of a summary}.
\newblock \emph{CoRR}, abs/2010.00490.

\bibitem[{Fabbri et~al.(2020)Fabbri, Ng, Wang, Nallapati, and
  Xiang}]{DBLP:journals/corr/abs-2004-11892}
Alexander~R. Fabbri, Patrick Ng, Zhiguo Wang, Ramesh Nallapati, and Bing Xiang.
  2020.
\newblock \href {http://arxiv.org/abs/2004.11892} {Template-based question
  generation from retrieved sentences for improved unsupervised question
  answering}.
\newblock \emph{CoRR}, abs/2004.11892.

\bibitem[{Haveliwala(2003)}]{haveliwala2003topic}
Taher~H Haveliwala. 2003.
\newblock Topic-sensitive pagerank: A context-sensitive ranking algorithm for
  web search.
\newblock \emph{IEEE transactions on knowledge and data engineering},
  15(4):784--796.

\bibitem[{Hearst(1997)}]{hearst1997text}
Marti~A Hearst. 1997.
\newblock Text tiling: Segmenting text into multi-paragraph subtopic passages.
\newblock \emph{Computational linguistics}, 23(1):33--64.

\bibitem[{Heilman and Smith(2010)}]{heilman2010good}
Michael Heilman and Noah~A Smith. 2010.
\newblock Good question! statistical ranking for question generation.
\newblock In \emph{Human Language Technologies: The 2010 Annual Conference of
  the North American Chapter of the Association for Computational Linguistics},
  pages 609--617.

\bibitem[{Huang et~al.(2021)Huang, Cao, Parulian, Ji, and
  Wang}]{huang2021efficient}
Luyang Huang, Shuyang Cao, Nikolaus Parulian, Heng Ji, and Lu~Wang. 2021.
\newblock Efficient attentions for long document summarization.
\newblock \emph{arXiv preprint arXiv:2104.02112}.

\bibitem[{Huang et~al.(2018)Huang, Hsieh, and Wang}]{huang2018automatic}
Tai-Chia Huang, Chia-Hsuan Hsieh, and Hei-Chia Wang. 2018.
\newblock Automatic meeting summarization and topic detection system.
\newblock \emph{Data Technologies and Applications}.

\bibitem[{Ishigaki et~al.(2020)Ishigaki, Huang, Takamura, Chen, and
  Okumura}]{ishigaki2020neural}
Tatsuya Ishigaki, Hen-Hsen Huang, Hiroya Takamura, Hsin-Hsi Chen, and Manabu
  Okumura. 2020.
\newblock Neural query-biased abstractive summarization using copying
  mechanism.
\newblock In \emph{European Conference on Information Retrieval}, pages
  174--181. Springer.

\bibitem[{Jacquenet et~al.(2019)Jacquenet, Bernard, and
  Largeron}]{jacquenet2019meeting}
Fran{\c{c}}ois Jacquenet, Marc Bernard, and Christine Largeron. 2019.
\newblock Meeting summarization, a challenge for deep learning.
\newblock In \emph{International Work-Conference on Artificial Neural
  Networks}, pages 644--655. Springer.

\bibitem[{Janin et~al.(2003)Janin, Baron, Edwards, Ellis, Gelbart, Morgan,
  Peskin, Pfau, Shriberg, Stolcke et~al.}]{janin2003icsi}
Adam Janin, Don Baron, Jane Edwards, Dan Ellis, David Gelbart, Nelson Morgan,
  Barbara Peskin, Thilo Pfau, Elizabeth Shriberg, Andreas Stolcke, et~al. 2003.
\newblock The icsi meeting corpus.
\newblock In \emph{2003 IEEE International Conference on Acoustics, Speech, and
  Signal Processing, 2003. Proceedings.(ICASSP'03).}, volume~1, pages I--I.
  IEEE.

\bibitem[{Jiang et~al.(2008)Jiang, Pei, Lin, Cheung, and Han}]{jiang2008mining}
Bin Jiang, Jian Pei, Xuemin Lin, David~W Cheung, and Jiawei Han. 2008.
\newblock Mining preferences from superior and inferior examples.
\newblock In \emph{Proceedings of the 14th ACM SIGKDD international conference
  on Knowledge discovery and data mining}, pages 390--398.

\bibitem[{Kim(2014)}]{DBLP:journals/corr/Kim14f}
Yoon Kim. 2014.
\newblock \href {http://arxiv.org/abs/1408.5882} {Convolutional neural networks
  for sentence classification}.
\newblock \emph{CoRR}, abs/1408.5882.

\bibitem[{Krippendorff(2011)}]{krippendorff2011computing}
Klaus Krippendorff. 2011.
\newblock Computing krippendorff's alpha-reliability.

\bibitem[{Krishna and Iyyer(2019)}]{krishna2019generating}
Kalpesh Krishna and Mohit Iyyer. 2019.
\newblock Generating question-answer hierarchies.
\newblock \emph{arXiv preprint arXiv:1906.02622}.

\bibitem[{Kry{\'s}ci{\'n}ski et~al.(2021)Kry{\'s}ci{\'n}ski, Rajani, Agarwal,
  Xiong, and Radev}]{kryscinski2021booksum}
Wojciech Kry{\'s}ci{\'n}ski, Nazneen Rajani, Divyansh Agarwal, Caiming Xiong,
  and Dragomir Radev. 2021.
\newblock Booksum: A collection of datasets for long-form narrative
  summarization.
\newblock \emph{arXiv preprint arXiv:2105.08209}.

\bibitem[{Kulkarni et~al.(2020)Kulkarni, Chammas, Zhu, Sha, and
  Ie}]{kulkarni2020aquamuse}
Sayali Kulkarni, Sheide Chammas, Wan Zhu, Fei Sha, and Eugene Ie. 2020.
\newblock Aquamuse: Automatically generating datasets for query-based
  multi-document summarization.
\newblock \emph{arXiv preprint arXiv:2010.12694}.

\bibitem[{Kulkarni et~al.(2021)Kulkarni, Chammas, Zhu, Sha, and
  Ie}]{kulkarni2021comsum}
Sayali Kulkarni, Sheide Chammas, Wan Zhu, Fei Sha, and Eugene Ie. 2021.
\newblock Comsum and sibert: A dataset and neural model for query-based
  multi-document summarization.
\newblock In \emph{International Conference on Document Analysis and
  Recognition}, pages 84--98. Springer.

\bibitem[{Kurland and Lee(2010)}]{kurland2010pagerank}
Oren Kurland and Lillian Lee. 2010.
\newblock Pagerank without hyperlinks: Structural reranking using links induced
  by language models.
\newblock \emph{ACM Transactions on Information Systems (TOIS)}, 28(4):1--38.

\bibitem[{Labutov et~al.(2015)Labutov, Basu, and Vanderwende}]{labutov2015deep}
Igor Labutov, Sumit Basu, and Lucy Vanderwende. 2015.
\newblock Deep questions without deep understanding.
\newblock In \emph{Proceedings of the 53rd Annual Meeting of the Association
  for Computational Linguistics and the 7th International Joint Conference on
  Natural Language Processing (Volume 1: Long Papers)}, pages 889--898.

\bibitem[{Lawson et~al.(2006)Lawson, Bodle, Houlette, and Haubner}]{Lawson2006}
Timothy~J. Lawson, James~H. Bodle, Melissa~A. Houlette, and Richard~R. Haubner.
  2006.
\newblock \href {https://doi.org/10.1207/s15328023top3301\_7} {Guiding
  questions enhance student learning from educational videos}.
\newblock \emph{Teaching of Psychology}, 33(1):31--33.

\bibitem[{Lawson et~al.(2007)Lawson, Bodle, and
  McDonough}]{lawson2007techniques}
Timothy~J Lawson, James~H Bodle, and Tracy~A McDonough. 2007.
\newblock Techniques for increasing student learning from educational videos:
  Notes versus guiding questions.
\newblock \emph{Teaching of Psychology}, 34(2):90--93.

\bibitem[{Lelkes et~al.(2021)Lelkes, Tran, and Yu}]{lelkes2021quiz}
Adam~D Lelkes, Vinh~Q Tran, and Cong Yu. 2021.
\newblock Quiz-style question generation for news stories.
\newblock In \emph{Proceedings of the Web Conference 2021}, pages 2501--2511.

\bibitem[{Li et~al.(2019)Li, Zhang, Ji, and Radke}]{li-etal-2019-keep}
Manling Li, Lingyu Zhang, Heng Ji, and Richard~J. Radke. 2019.
\newblock \href {https://doi.org/10.18653/v1/P19-1210} {Keep meeting summaries
  on topic: Abstractive multi-modal meeting summarization}.
\newblock In \emph{Proceedings of the 57th Annual Meeting of the Association
  for Computational Linguistics}, pages 2190--2196, Florence, Italy.
  Association for Computational Linguistics.

\bibitem[{Lin(2004)}]{lin-2004-rouge}
Chin-Yew Lin. 2004.
\newblock \href {https://aclanthology.org/W04-1013} {{ROUGE}: A package for
  automatic evaluation of summaries}.
\newblock In \emph{Text Summarization Branches Out}, pages 74--81, Barcelona,
  Spain. Association for Computational Linguistics.

\bibitem[{Lindberg et~al.(2013)Lindberg, Popowich, Nesbit, and
  Winne}]{lindberg2013generating}
David Lindberg, Fred Popowich, John Nesbit, and Phil Winne. 2013.
\newblock Generating natural language questions to support learning on-line.
\newblock In \emph{Proceedings of the 14th European Workshop on Natural
  Language Generation}, pages 105--114.

\bibitem[{Litvak and Vanetik(2017)}]{litvak-vanetik-2017-query}
Marina Litvak and Natalia Vanetik. 2017.
\newblock \href {https://doi.org/10.18653/v1/W17-1004} {Query-based
  summarization using {MDL} principle}.
\newblock In \emph{Proceedings of the {M}ulti{L}ing 2017 Workshop on
  Summarization and Summary Evaluation Across Source Types and Genres}, pages
  22--31, Valencia, Spain. Association for Computational Linguistics.

\bibitem[{Liu et~al.(2019)Liu, Yuan, Butler, Carvajal, Li, Ta, and
  Weng}]{liu2019dquest}
Cong Liu, Chi Yuan, Alex~M Butler, Richard~D Carvajal, Ziran~Ryan Li, Casey~N
  Ta, and Chunhua Weng. 2019.
\newblock Dquest: dynamic questionnaire for search of clinical trials.
\newblock \emph{Journal of the American Medical Informatics Association},
  26(11):1333--1343.

\bibitem[{Liu and Lapata(2019)}]{liu2019text}
Yang Liu and Mirella Lapata. 2019.
\newblock Text summarization with pretrained encoders.
\newblock \emph{arXiv preprint arXiv:1908.08345}.

\bibitem[{Manakul and Gales(2021)}]{manakul2021long}
Potsawee Manakul and Mark Gales. 2021.
\newblock Long-span summarization via local attention and content selection.
\newblock In \emph{Proceedings of the 59th Annual Meeting of the Association
  for Computational Linguistics and the 11th International Joint Conference on
  Natural Language Processing (Volume 1: Long Papers)}, pages 6026--6041.

\bibitem[{Maynez et~al.(2020)Maynez, Narayan, Bohnet, and
  McDonald}]{maynez-etal-2020-faithfulness}
Joshua Maynez, Shashi Narayan, Bernd Bohnet, and Ryan McDonald. 2020.
\newblock \href {https://doi.org/10.18653/v1/2020.acl-main.173} {On
  faithfulness and factuality in abstractive summarization}.
\newblock In \emph{Proceedings of the 58th Annual Meeting of the Association
  for Computational Linguistics}, pages 1906--1919, Online. Association for
  Computational Linguistics.

\bibitem[{Mazidi and Nielsen(2014)}]{mazidi-nielsen-2014-linguistic}
Karen Mazidi and Rodney~D. Nielsen. 2014.
\newblock \href {https://doi.org/10.3115/v1/P14-2053} {Linguistic
  considerations in automatic question generation}.
\newblock In \emph{Proceedings of the 52nd Annual Meeting of the Association
  for Computational Linguistics (Volume 2: Short Papers)}, pages 321--326,
  Baltimore, Maryland. Association for Computational Linguistics.

\bibitem[{Mehdad et~al.(2013)Mehdad, Carenini, Tompa, and
  Ng}]{mehdad2013abstractive}
Yashar Mehdad, Giuseppe Carenini, Frank Tompa, and Raymond Ng. 2013.
\newblock Abstractive meeting summarization with entailment and fusion.
\newblock In \emph{Proceedings of the 14th European Workshop on Natural
  Language Generation}, pages 136--146.

\bibitem[{Mostow and Chen(2009)}]{mostow2009generating}
Jack Mostow and Wei Chen. 2009.
\newblock Generating instruction automatically for the reading strategy of
  self-questioning.
\newblock In \emph{AIED}, pages 465--472.

\bibitem[{Murray et~al.(2010)Murray, Carenini, and Ng}]{murray2010generating}
Gabriel Murray, Giuseppe Carenini, and Raymond Ng. 2010.
\newblock Generating and validating abstracts of meeting conversations: a user
  study.
\newblock In \emph{Proceedings of the 6th International Natural Language
  Generation Conference}.

\bibitem[{Nallapati et~al.(2017)Nallapati, Zhai, and
  Zhou}]{nallapati2017summarunner}
Ramesh Nallapati, Feifei Zhai, and Bowen Zhou. 2017.
\newblock Summarunner: A recurrent neural network based sequence model for
  extractive summarization of documents.
\newblock In \emph{Thirty-First AAAI Conference on Artificial Intelligence}.

\bibitem[{Nema et~al.(2017)Nema, Khapra, Laha, and
  Ravindran}]{nema-etal-2017-diversity}
Preksha Nema, Mitesh~M. Khapra, Anirban Laha, and Balaraman Ravindran. 2017.
\newblock \href {https://doi.org/10.18653/v1/P17-1098} {Diversity driven
  attention model for query-based abstractive summarization}.
\newblock In \emph{Proceedings of the 55th Annual Meeting of the Association
  for Computational Linguistics (Volume 1: Long Papers)}, pages 1063--1072,
  Vancouver, Canada. Association for Computational Linguistics.

\bibitem[{Oya et~al.(2014)Oya, Mehdad, Carenini, and
  Ng}]{oya-etal-2014-template}
Tatsuro Oya, Yashar Mehdad, Giuseppe Carenini, and Raymond Ng. 2014.
\newblock \href {https://doi.org/10.3115/v1/W14-4407} {A template-based
  abstractive meeting summarization: Leveraging summary and source text
  relationships}.
\newblock In \emph{Proceedings of the 8th International Natural Language
  Generation Conference ({INLG})}, pages 45--53, Philadelphia, Pennsylvania,
  U.S.A. Association for Computational Linguistics.

\bibitem[{Papineni et~al.(2002)Papineni, Roukos, Ward, and
  Zhu}]{papineni2002bleu}
Kishore Papineni, Salim Roukos, Todd Ward, and Wei-Jing Zhu. 2002.
\newblock Bleu: a method for automatic evaluation of machine translation.
\newblock In \emph{Proceedings of the 40th annual meeting of the Association
  for Computational Linguistics}, pages 311--318.

\bibitem[{Pasunuru et~al.(2021)Pasunuru, Celikyilmaz, Galley, Xiong, Zhang,
  Bansal, and Gao}]{pasunuru2021data}
Ramakanth Pasunuru, Asli Celikyilmaz, Michel Galley, Chenyan Xiong, Yizhe
  Zhang, Mohit Bansal, and Jianfeng Gao. 2021.
\newblock Data augmentation for abstractive query-focused multi-document
  summarization.
\newblock In \emph{Proceedings of the AAAI Conference on Artificial
  Intelligence}, volume~35, pages 13666--13674.

\bibitem[{Rajpurkar et~al.(2016)Rajpurkar, Zhang, Lopyrev, and
  Liang}]{rajpurkar2016squad}
Pranav Rajpurkar, Jian Zhang, Konstantin Lopyrev, and Percy Liang. 2016.
\newblock Squad: 100,000+ questions for machine comprehension of text.
\newblock \emph{arXiv preprint arXiv:1606.05250}.

\bibitem[{Reddy et~al.(2019)Reddy, Chen, and Manning}]{reddy2019coqa}
Siva Reddy, Danqi Chen, and Christopher~D Manning. 2019.
\newblock Coqa: A conversational question answering challenge.
\newblock \emph{Transactions of the Association for Computational Linguistics},
  7:249--266.

\bibitem[{Reimers and Gurevych(2019)}]{DBLP:journals/corr/abs-1908-10084}
Nils Reimers and Iryna Gurevych. 2019.
\newblock \href {http://arxiv.org/abs/1908.10084} {Sentence-bert: Sentence
  embeddings using siamese bert-networks}.
\newblock \emph{CoRR}, abs/1908.10084.

\bibitem[{Rokach and Kisilevich(2012)}]{rokach2012initial}
Lior Rokach and Slava Kisilevich. 2012.
\newblock Initial profile generation in recommender systems using pairwise
  comparison.
\newblock \emph{IEEE Transactions on Systems, Man, and Cybernetics, Part C
  (Applications and Reviews)}, 42(6):1854--1859.

\bibitem[{Rus et~al.(2010)Rus, Wyse, Piwek, Lintean, Stoyanchev, and
  Moldovan}]{rus2010first}
Vasile Rus, Brendan Wyse, Paul Piwek, Mihai Lintean, Svetlana Stoyanchev, and
  Cristian Moldovan. 2010.
\newblock The first question generation shared task evaluation challenge.

\bibitem[{Rush et~al.(2015)Rush, Chopra, and Weston}]{rush2015neural}
Alexander~M Rush, Sumit Chopra, and Jason Weston. 2015.
\newblock A neural attention model for abstractive sentence summarization.
\newblock \emph{arXiv preprint arXiv:1509.00685}.

\bibitem[{Sanh et~al.(2019)Sanh, Debut, Chaumond, and
  Wolf}]{sanh2019distilbert}
Victor Sanh, Lysandre Debut, Julien Chaumond, and Thomas Wolf. 2019.
\newblock Distilbert, a distilled version of bert: smaller, faster, cheaper and
  lighter.
\newblock \emph{arXiv preprint arXiv:1910.01108}.

\bibitem[{See et~al.(2017)See, Liu, and Manning}]{DBLP:journals/corr/SeeLM17}
Abigail See, Peter~J. Liu, and Christopher~D. Manning. 2017.
\newblock \href {http://arxiv.org/abs/1704.04368} {Get to the point:
  Summarization with pointer-generator networks}.
\newblock \emph{CoRR}, abs/1704.04368.

\bibitem[{Sehikh et~al.(2017)Sehikh, Fohr, and Illina}]{sehikh2017topic}
Imran Sehikh, Dominique Fohr, and Irina Illina. 2017.
\newblock Topic segmentation in asr transcripts using bidirectional rnns for
  change detection.
\newblock In \emph{2017 IEEE Automatic Speech Recognition and Understanding
  Workshop (ASRU)}, pages 512--518. IEEE.

\bibitem[{Sepliarskaia et~al.(2018)Sepliarskaia, Kiseleva, Radlinski, and
  de~Rijke}]{sepliarskaia2018preference}
Anna Sepliarskaia, Julia Kiseleva, Filip Radlinski, and Maarten de~Rijke. 2018.
\newblock Preference elicitation as an optimization problem.
\newblock In \emph{Proceedings of the 12th ACM Conference on Recommender
  Systems}, pages 172--180.

\bibitem[{Shang et~al.(2018)Shang, Ding, Zhang, Tixier, Meladianos,
  Vazirgiannis, and Lorr{\'e}}]{shang-etal-2018-unsupervised}
Guokan Shang, Wensi Ding, Zekun Zhang, Antoine Tixier, Polykarpos Meladianos,
  Michalis Vazirgiannis, and Jean-Pierre Lorr{\'e}. 2018.
\newblock \href {https://doi.org/10.18653/v1/P18-1062} {Unsupervised
  abstractive meeting summarization with multi-sentence compression and
  budgeted submodular maximization}.
\newblock In \emph{Proceedings of the 56th Annual Meeting of the Association
  for Computational Linguistics (Volume 1: Long Papers)}, pages 664--674,
  Melbourne, Australia. Association for Computational Linguistics.

\bibitem[{Singhal et~al.(2020)Singhal, Khatter, Tejaswini, and
  Jayashree}]{singhal2020abstractive}
Daksha Singhal, Kavya Khatter, A~Tejaswini, and R~Jayashree. 2020.
\newblock Abstractive summarization of meeting conversations.
\newblock In \emph{2020 IEEE International Conference for Innovation in
  Technology (INOCON)}, pages 1--4. IEEE.

\bibitem[{Solbiati et~al.(2021)Solbiati, Heffernan, Damaskinos, Poddar, Modi,
  and Cali}]{solbiati2021unsupervised}
Alessandro Solbiati, Kevin Heffernan, Georgios Damaskinos, Shivani Poddar,
  Shubham Modi, and Jacques Cali. 2021.
\newblock Unsupervised topic segmentation of meetings with bert embeddings.
\newblock \emph{arXiv preprint arXiv:2106.12978}.

\bibitem[{ter Hoeve et~al.(2020)ter Hoeve, Kiseleva, and
  de~Rijke}]{ter2020makes}
Maartje ter Hoeve, Julia Kiseleva, and Maarten de~Rijke. 2020.
\newblock What makes a good summary? reconsidering the focus of automatic
  summarization.

\bibitem[{ter Hoeve et~al.(2022)ter Hoeve, Kiseleva, and
  de~Rijke}]{ter2022summarization}
Maartje ter Hoeve, Julia Kiseleva, and Maarten de~Rijke. 2022.
\newblock Summarization with graphical elements.
\newblock \emph{arXiv preprint arXiv:2204.07551}.

\bibitem[{Vaswani et~al.(2017)Vaswani, Shazeer, Parmar, Uszkoreit, Jones,
  Gomez, Kaiser, and Polosukhin}]{vaswani2017attention}
Ashish Vaswani, Noam Shazeer, Niki Parmar, Jakob Uszkoreit, Llion Jones,
  Aidan~N Gomez, {\L}ukasz Kaiser, and Illia Polosukhin. 2017.
\newblock Attention is all you need.
\newblock In \emph{Advances in Neural Information Processing Systems}, pages
  5998--6008.

\bibitem[{Wallach(2004)}]{wallach2004conditional}
Hanna~M Wallach. 2004.
\newblock Conditional random fields: An introduction.
\newblock \emph{Technical Reports (CIS)}, page~22.

\bibitem[{Wang et~al.(2020)Wang, Cho, and Lewis}]{wang-etal-2020-asking}
Alex Wang, Kyunghyun Cho, and Mike Lewis. 2020.
\newblock \href {https://doi.org/10.18653/v1/2020.acl-main.450} {Asking and
  answering questions to evaluate the factual consistency of summaries}.
\newblock In \emph{Proceedings of the 58th Annual Meeting of the Association
  for Computational Linguistics}, pages 5008--5020, Online. Association for
  Computational Linguistics.

\bibitem[{Wang and Cardie(2013)}]{wang-cardie-2013-domain}
Lu~Wang and Claire Cardie. 2013.
\newblock \href {https://www.aclweb.org/anthology/P13-1137} {Domain-independent
  abstract generation for focused meeting summarization}.
\newblock In \emph{Proceedings of the 51st Annual Meeting of the Association
  for Computational Linguistics (Volume 1: Long Papers)}, pages 1395--1405,
  Sofia, Bulgaria. Association for Computational Linguistics.

\bibitem[{Zhang et~al.(2020)Zhang, Zhao, Saleh, and Liu}]{zhang2020pegasus}
Jingqing Zhang, Yao Zhao, Mohammad Saleh, and Peter Liu. 2020.
\newblock Pegasus: Pre-training with extracted gap-sentences for abstractive
  summarization.
\newblock In \emph{International Conference on Machine Learning}, pages
  11328--11339. PMLR.

\bibitem[{Zhang and Zhou(2019)}]{zhang2019topic}
Leilan Zhang and Qiang Zhou. 2019.
\newblock Topic segmentation for dialogue stream.
\newblock In \emph{2019 Asia-Pacific Signal and Information Processing
  Association Annual Summit and Conference (APSIPA ASC)}, pages 1036--1043.
  IEEE.

\bibitem[{Zhao et~al.(2019)Zhao, Pan, Fan, Liu, Li, Yang, and
  Cai}]{zhao2019abstractive}
Zhou Zhao, Haojie Pan, Changjie Fan, Yan Liu, Linlin Li, Min Yang, and Deng
  Cai. 2019.
\newblock Abstractive meeting summarization via hierarchical adaptive segmental
  network learning.
\newblock In \emph{The World Wide Web Conference}, pages 3455--3461.

\bibitem[{Zhong et~al.(2021)Zhong, Yin, Yu, Zaidi, Mutuma, Jha, Awadallah,
  Celikyilmaz, Liu, Qiu, and Radev}]{zhong2021qmsum}
Ming Zhong, Da~Yin, Tao Yu, Ahmad Zaidi, Mutethia Mutuma, Rahul Jha,
  Ahmed~Hassan Awadallah, Asli Celikyilmaz, Yang Liu, Xipeng Qiu, and Dragomir
  Radev. 2021.
\newblock \href {https://doi.org/10.18653/v1/2021.naacl-main.472} {{QMS}um: A
  new benchmark for query-based multi-domain meeting summarization}.
\newblock In \emph{Proceedings of the 2021 Conference of the North American
  Chapter of the Association for Computational Linguistics: Human Language
  Technologies}, pages 5905--5921, Online. Association for Computational
  Linguistics.

\bibitem[{Zhu et~al.(2020)Zhu, Xu, Zeng, and
  Huang}]{zhu-etal-2020-hierarchical}
Chenguang Zhu, Ruochen Xu, Michael Zeng, and Xuedong Huang. 2020.
\newblock \href {https://www.aclweb.org/anthology/2020.findings-emnlp.19} {A
  hierarchical network for abstractive meeting summarization with cross-domain
  pretraining}.
\newblock In \emph{Findings of the Association for Computational Linguistics:
  EMNLP 2020}, pages 194--203, Online. Association for Computational
  Linguistics.

\end{thebibliography}
\bibliographystyle{acl_natbib}

\end{document}